\documentclass[10pt,conference,compsocconf]{IEEEtran}
\IEEEoverridecommandlockouts
\usepackage{cite}
\usepackage{amsmath,amssymb,amsfonts}
\usepackage{graphicx}

\usepackage{subcaption}
\usepackage{textcomp}
\usepackage{adjustbox}
\usepackage{listings}
\usepackage{xcolor}
\usepackage{comment}
\usepackage{lipsum}
\usepackage{tabularx}
\usepackage[ruled,linesnumbered]{algorithm2e} 
\usepackage{algorithmic} 
\usepackage{multirow}
\usepackage{graphics}
\usepackage{booktabs}
\usepackage{tablefootnote}
\usepackage{xcolor}

\def\IEEEbibitemsep{0pt plus .5pt}

\usepackage{lipsum} 

\usepackage{amsthm}

\theoremstyle{definition}
\newtheorem{myobj}{Objective Function}

\usepackage{textcomp}
\def\BibTeX{{\rm B\kern-.05em{\sc i\kern-.025em b}\kern-.08em
		T\kern-.1667em\lower.7ex\hbox{E}\kern-.125emX}}
		
\newlength\mylenin
\newcommand\myinput[1]{%
\settowidth\mylenin{\KwIn{}}%
\setlength\hangindent{\mylenin}%
\hspace*{\mylenin}#1\\}

\let\oldnl\nl
\newcommand{\nonl}{\renewcommand{\nl}{\let\nl\oldnl}}

\newlength\mylenout
\newcommand\myoutput[1]{%
\settowidth\mylenout{\KwOut{}}%
\setlength\hangindent{\mylenout}%
\hspace*{\mylenout}#1\\}



\usepackage{afterpage}
\newlength{\oldtextfloatsep}
		
\begin{document}

	\title{
	\vspace{-1cm}
		f-CN$\text{N}^{\text{x}}$: A Toolflow for Mapping Multi-CNN Applications on FPGAs 
	\vspace{-1cm}
	}
		


	\author{\IEEEauthorblockN{Stylianos I. Venieris}
		\IEEEauthorblockA{{Department of Electrical and Electronic Engineering} \\
			{Imperial College London}\\
			Email: stylianos.venieris10@imperial.ac.uk 
			\vspace{-0.5cm}}
		\and
		\IEEEauthorblockN{Christos-Savvas Bouganis}
		\IEEEauthorblockA{{Department of Electrical and Electronic Engineering} \\
			{Imperial College London}\\
			Email: christos-savvas.bouganis@imperial.ac.uk 
			\vspace{-0.5cm}}
	}
	
	\maketitle

	\begin{abstract}
	
		The predictive power of Convolutional Neural Networks (CNNs) has been an integral factor for emerging latency-sensitive applications, such as autonomous drones and vehicles. Such systems employ multiple CNNs, each one trained for a particular task. 
		The efficient mapping of multiple CNNs on a single FPGA device is a challenging task as the allocation of compute resources and external memory bandwidth needs to be optimised at design time. 
		This paper proposes f-CN$\text{N}^{\text{x}}$, an automated toolflow for the optimised mapping of multiple CNNs on FPGAs, comprising a novel multi-CNN hardware architecture together with an automated design space exploration method that considers the user-specified performance requirements for each model to allocate compute resources and generate a synthesisable accelerator. 
		Moreover, f-CNN$^{\text{x}}$ employs a novel scheduling algorithm 
		that alleviates the limitations of the memory bandwidth 
		contention between CNNs and sustains the high utilisation of the architecture. 
		Experimental evaluation shows that f-CN$\text{N}^{\text{x}}$'s 
		designs outperform contention-unaware FPGA mappings by up to 50\% and deliver up to 6.8x higher performance-per-Watt over highly optimised GPU designs for multi-CNN systems.
		
	\end{abstract}
	
%
	\vspace{-0.15cm}
	\section{Introduction}
	\vspace{-0.15cm}
	Over the last decade, Convolutional Neural Network (CNN) models have substantially improved the state-of-the-art performance in several Artificial Intelligence (AI) tasks. This property has made CNNs an enabling technology for novel systems in both embedded and cloud applications. On the one side of the spectrum, autonomous robots and vehicles is an emerging field that has gathered wide interest from both the academic \cite{Chen2015} and industrial \cite{nvidia_2016} communities due to its potential societal and economic effects. On the other end, data centre-based analytics that employ CNNs are becoming a widespread operational model to serve a large pool of clients.
%
	
	
	
	
	\vspace{-0.1cm}
	Both embedded and cloud AI systems rely their operation on multiple CNNs. In {latency-critical}, vision-centric autonomous systems, perception is largely based on highly accurate and reliable computer vision tasks, such as object detection \cite{Ren_2017} and semantic segmentation \cite{Badrinarayanan_2017}. 
	Similarly, cloud-based systems have to cope with servicing 
	a wide range of concurrent CNN applications, from bioinformatics to visual search \cite{Caulfield2016}, {with stringent response-time demands}. 
	In such scenarios, a dedicated model is trained for each particular task, leading to the parallel execution of several CNNs on the same target platform. {Moreover, the latency-sensitive nature of modern applications prohibits the use of batch processing.} As a result, in both emerging embedded and cloud applications there is a requirement for the {latency-driven} mapping of multiple CNNs on the computing platform of the target system. 
	
	\vspace{-0.1cm}
	Currently, the conventional computing infrastructure of complex autonomous systems and data centres 
	comprises CPUs and GPUs
	, which are able to provide high processing speed 
	at the expense of high power consumption. A potential alternative platform that can offer both the flexibility and performance that is required by modern CNNs at a lower power envelope are the FPGAs. 
	{In the space of multi-CNN systems, FPGAs offer unique optimisation opportunities due to the possibility of fine-grained allocation of resources, which is not offered by other platforms.}
	However, until now, CNN implementations, including FPGA-based accelerators {\cite{Venieris_2017b,Yufei_Ma_2017,sv2018csur}}, are typically designed and optimised for scenarios where a single model is running for an extensive period of time, while the multiple CNNs setting has remained unexplored.
	\vspace{-0.1cm}
	{In this paper, we propose f-CN$\text{N}^{\text{x}}$, an automated framework that maps multiple CNNs on a target FPGA, by taking into account the application-level required performance for each model and the available hardware resources, in order to generate an optimised multi-CNN architecture. The proposed framework exploits the structure of CNN workloads and the fine-grained control over resource allocation of FPGAs to yield latency-optimised designs that overcome the limitations of other parallel platforms targeting multiple CNNs.} 
	This paper makes the following key contributions:
	\vspace{-0.1cm}
	\begin{itemize}
	\vspace{-0.125cm}
		\item A novel architecture for the parallel execution of multiple CNNs on a single FPGA. {The proposed architecture is parametrised to allow the fine-grained allocation of resources among CNNs and the deterministic scheduling of external memory transfers to minimise memory contention. This parametrisation enables us to explore the design space of a wide range of resource and bandwidth allocations.}
		
		\vspace{-0.125cm}
		\item A novel design space exploration algorithm for efficiently traversing the large design space. The proposed algorithm co-optimises the mapping of multiple CNNs on the target FPGA and incorporates the application-level importance of each model by means of multiobjective cost functions in order to guide the design space exploration to the {\color{black}optimum} design points. Moreover, a scheduling algorithm is proposed for the optimised sharing of the external memory bandwidth.
		
		\vspace{-0.125cm}
		\item {The f-CN$\text{N}^{\text{x}}$ automated toolflow for mapping multiple CNNs on a particular FPGA-based platform, 
		taking as input a target set of CNNs in a high-level description, performing fast design space exploration and generating a synthesisable Vivado HLS hardware design. 
		}
	\end{itemize}
	
	\vspace{-0.2cm}
	\noindent
	To the best of our knowledge, this work addresses for the first time in the literature the mapping of multiple CNNs.
	
	
	
	

	\vspace{-0.2cm}
	\section{Multiple CNNs on Reconfigurable Logic}
	\vspace{-1mm}
	\subsection{Background on Multi-CNN Systems}
	\vspace{-1mm}
	Multi-CNN systems employ a number of models, with each one trained for a different 
	task. In the embedded space, drones and self-driving cars run a variety of concurrent tasks, such as navigation and obstacle avoidance \cite{Smolyanskiy_2017}
	. In the cloud, services are increasingly heterogeneous, with diverse workloads executed concurrently for a large number of users \cite{Caulfield2016}. 
	Nevertheless, mapping multiple CNNs on a computing platform poses a challenge. With each model targeting a different task, the performance {\color{black}constraints}, such as required throughput and latency, vary accordingly. {Moreover, in resource-constrained setups, 
	multiple CNNs compete for the same pool of computational and memory resources. As a result, the mapping of multiple CNNs is a high-dimensional design problem that encompasses both the performance needs of each model and the resource constraints of the target platform. 
	}
	
%
	
	
	\vspace{-3mm}
	\subsection{Opportunities and Challenges in Mapping Multiple CNNs}
	\vspace{-1mm}
	\label{sec:opps_section}
	CNNs comprise a sequence of layers, organised as a feature extractor and a classifier stage. 
	With the feature extractor dominating the computational cost and fully-connected layers limited in recent state-of-the-art models {\cite{He_2016,Howard2017mobilenet,Zoph2018cvpr}}, this work focuses on the feature extractor. In the context of multiple CNNs, their characteristic structure presents opportunities for performance optimisation. 
	The dataflow of a CNN consists of 
	a feed-forward topology which can be modelled as a directed acyclic graph with one node per layer. Under this model, the dependencies between nodes 
	and the workload of each node, including the ops/input, storage and memory bandwidth for weights and feature maps, are known a priori based on each layer's type and configuration. 
	This prior knowledge of compute and memory requirements enables 1) optimising at compile time the on-chip resource allocation between multiple CNNs and 2) generating an optimised static schedule for sharing the bandwidth to sustain high hardware utilisation. 
	
	\vspace{-1mm}
	To exploit effectively these CNN-specific opportunities, a fine-grained control over the customisation of the hardware is required. Fine-grained parametrisation would allow tailoring the allocation of on-chip resources to the potentially different performance needs of the CNNs. At the same time, control over the shared off-chip memory bandwidth would enable deriving a schedule that sustains a high utilisation of the architecture. Nevertheless, such a fine granularity leads to a large number of design parameters even for a single CNN. By scaling the problem to multiple models, the space of possible designs becomes combinatorially large. Thus, the complexity of mapping multiple CNNs on FPGAs necessitates a principled methodology in order to generate optimised designs.
	

	%
	
	\vspace{-2mm}
	\section{Proposed Framework}
	\vspace{-1mm}
	{\color{black}A high-level description of f-CN$\text{N}^{\text{x}}$'s flow is as follows.} The deep learning specialist provides the set of CNNs in Caffe\footnote{http://caffe.berkeleyvision.org/} format, together with a target performance for each model, and the resources of the target FPGA platform. The Caffe descriptions are translated to a dataflow representation with one node per layer and passed to the design space exploration (DSE). The DSE employs a Synchronous Dataflow \cite{Lee_1987} model of the multi-CNN hardware architecture and a memory scheduling policy to traverse the design space and optimise a multiobjective criterion that captures the user-specified performance for each CNN. After the highest performing design point has been selected, f-CN$\text{N}^{\text{x}}$ generates synthesisable Vivado HLS code, which is compiled by the vendor's toolchain.

	\begin{figure}[t]
	\vspace{-2mm}
		\centering
		\includegraphics[trim={0cm 4cm 11cm 2cm},clip,width=0.5\textwidth]{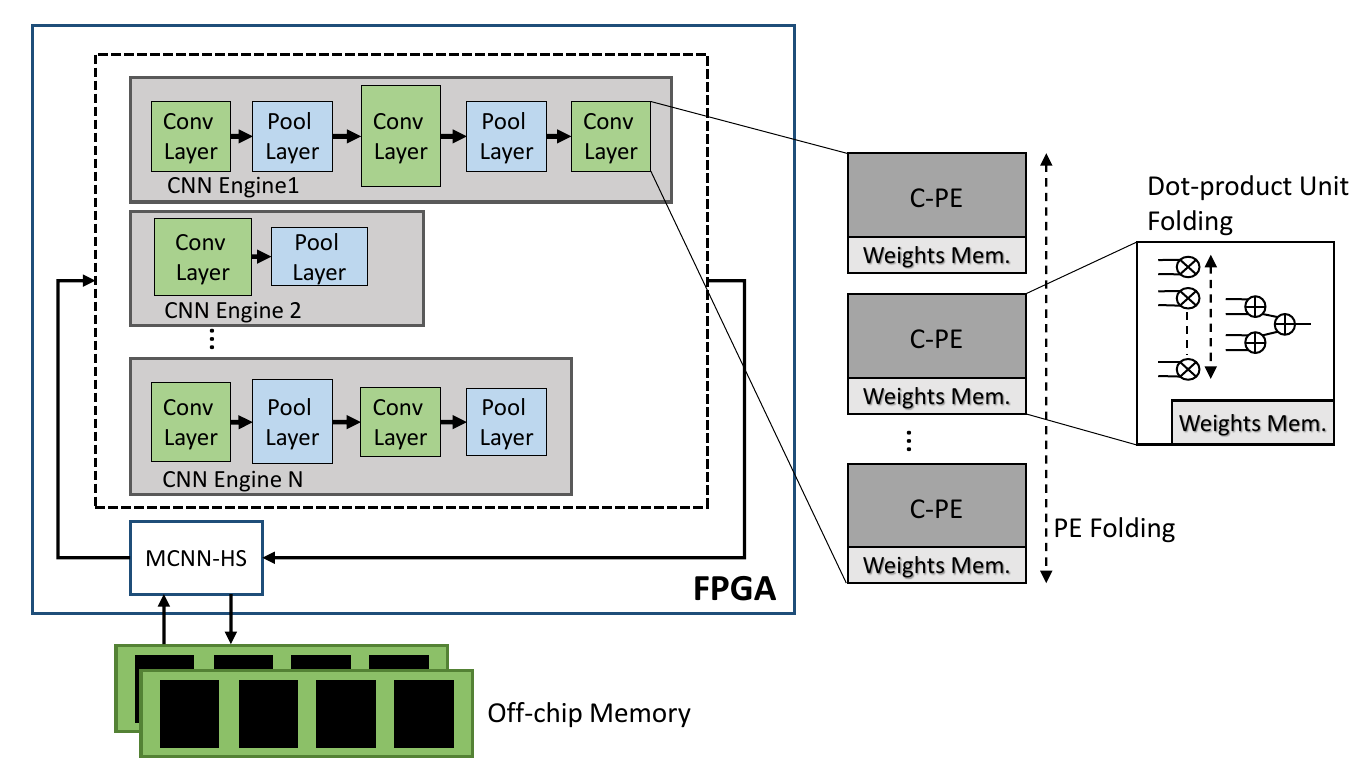}
		\vspace{-0.5cm}
		\caption{Parallel architecture for multiple CNNs.}
		\label{fig:multicnn_arch}
		\vspace{-0.91cm}
	\end{figure}
	
	\vspace{-3mm}
	\subsection{Architecture}
	\label{arch_sec}
	\vspace{-0.5mm}
	
	Fig. \ref{fig:multicnn_arch} shows the proposed multi-CNN architecture consisting of two components: a number of 
	heterogeneous CNN engines and a multi-CNN hardware scheduler (\texttt{MCNN-HS}). Instead of scheduling the target set of CNNs sequentially over a fixed accelerator, 
	the strategy of our framework is to generate one dedicated 
	engine per CNN, customised to its workload and performance needs, allowing the concurrent execution of all models in an efficient way. The \texttt{MCNN-HS} module 
	allocates the off-chip memory bandwidth to the CNN engines, with a static schedule as determined during the design space exploration. The scheduling of off-chip memory transactions and the design of \texttt{MCNN-HS} are detailed in Sec. \ref{sec:sched} and \ref{sec:mcnnhs} respectively.
	
	\vspace{-0.15cm}
	\textbf{CNN Engine.}
	The 
	hardware structure for each CNN engine can be either a core that processes each layer sequentially in a tiled manner (e.g. 
	a matrix multiplication unit or a systolic array) or a 
	streaming architecture. 
	In the first case, the engines would have a fixed hardware template with 
	customisable tile sizes. 
	In the latter case, a streaming design would be parametrised with respect to the instantiated stages, their interconnections and the resource allocation among them. In this work, the streaming paradigm is adopted, to obtain a finer grain of control over the structure of each individual CNN engine. 
	Each engine consists of a coarse pipeline of heterogeneous hardware stages, 
	with each stage parametrised with respect to its parallelism-resource trade-off. The pipeline for each CNN can have a different structure, with a customisable sequence of stages based on the topology and the computational needs of the corresponding CNN \mbox{(Fig. \ref{fig:multicnn_arch})}. Overall, the CNN engines operate under a data-driven scheme so that each stage computes whenever data arrive at its input.
	
	\vspace{-0.15cm}
	The hardware stages are composed of modules for the convolutional, pooling and nonlinear layers. In the convolutional layer, we exploit the parallelism with respect to its outputs by tunably unrolling and instantiating one convolution processing element (C-PE) per output feature map, with the input feature maps processed in a pipelined manner. The output feature maps are parametrised to be folded, as shown in \mbox{Fig. \ref{fig:multicnn_arch}}, so that C-PEs can be time-shared within a layer. Moreover, the dot-product circuit inside each C-PE can be tunably scaled (Fig. \ref{fig:multicnn_arch}), from a single multiply-accumulate operator up to a fully parallel 
	multiplier array with an adder tree. Pooling and nonlinear stages also have a tunable number of PEs, while operator-level folding can be applied on max and average units of pooling PEs. Under this parametrisation, each hardware stage has a tunable number of PEs, $N_{\text{PE}} \in [1,N_{\text{out}}]$, where $N_{\text{out}}$ is the maximum number of output feature maps it has to process, and a tunable number of 
	operators, $N_{\text{op}} \in [1,K^2]$, where $K$ is the filter or pooling size depending on the type of layer, and both can be optimised as dictated by the workload and the performance requirements of the particular CNN.
	
	\vspace{-0.15cm}
	With modern CNNs requiring an excessive amount of memory for their trained weights even for a single \mbox{layer \cite{He_2016},} we allow for the further folding of convolutional layers with respect to their inputs. Layers that exceed the on-chip storage of the target FPGA are tunably folded with respect to their input feature maps and the associated weights with a factor of \mbox{$f_{\text{in}} \in [1,N_{\text{in}}]$} which determines the tile size, where $N_{\text{in}}$ is the number of input feature maps. This approach enables the on-chip compute and memory resources allocated for a convolutional layer to be time-multiplexed 
	and the on-chip storage requirements to be accommodated by the target device.

\vspace{-0.2cm}
	\textbf{CNN Partitioning and Subgraphs.}
	\label{sec:partition}
	The large depth and amount of weights often prohibit the direct mapping of each individual CNN to hardware.
	To sustain the utilisation of the architecture, 
	we partition each CNN into subgraphs. The adopted partitioning scheme allows the partitioning along \mbox{1) the} depth of the model and 2) the input feature maps of each convolutional layer, and requires each subgraph to contain at least one convolutional layer. 
	With this formulation, the structure of each CNN engine is derived so that its datapath can execute all the subgraphs of the corresponding CNN.
	The partition points and the datapath for each engine are selected during the proposed design space exploration, described in Sec. \ref{sec:dse}. Given a set of partitioned CNNs, the compute and memory requirements of each subgraph are known at compile time, based on the subgraph's layers. As a result, the scheduling of the subgraphs on the corresponding engine as well as the memory transactions of the overall multi-CNN architecture can be  statically optimised at compile time.

	\vspace{-3mm}
	\subsection{Design Space Exploration}
	\vspace{-1mm}
	\label{sec:dse}
	Given a set of CNNs, the design space of possible mappings is formed by the free parameters of the architecture. These include 1) the partition points of each CNN, \mbox{2) the} structure of each CNN engine, including the number and type of hardware stages and the connections between stages, 3) the compile-time configurable folding parameters $\left<N_{\text{PE}}, N_{\text{op}}, f_{\text{in}}\right>$ of each stage,	and 4) the external memory bandwidth schedule. By defining such a large parameter space, our proposed framework trades off the capability of very fine-grained customisation that enables exploring a wide range of optimisations, at the cost of a combinatorial space of possible mappings. To capture each design point analytically and navigate efficiently the design space, we employ a Synchronous Dataflow (SDF) model \cite{Lee_1987} which considers the configuration of each candidate design to estimate performance, on-chip resource consumption and external memory bandwidth requirements.

	\vspace{-0.2cm}
	\textbf{Performance Model.}	
	Using the methodology described in \cite{Venieris_2016}, we develop an SDF model for the multi-CNN architecture. We model each CNN engine as an SDF graph $G_{\text{CE}}$$=$$(V,E)$, with each node $v \in V$ representing a hardware stage. The configuration of each stage in the CNN engine is captured with a tuple of the form $\left<N_{\text{PE}}, N_{\text{op}}, f_{\text{in}}, T\right>$, with $N_{\text{PE}}$, $N_{\text{op}}$ and $f_{\text{in}}$ as defined in Sec. \ref{arch_sec} and $T$ the type of module. In this setting, each stage has a consumption rate of $N_{\text{PE}}N_{\text{op}}$ elements/cycle and the CNN engine is equivalently represented with a topology matrix $\boldsymbol{\Gamma} \in \mathbb{R}^{|E|\times|V|}$ with $\boldsymbol{\Gamma}(e,v)$ holding the processing rate of node $v$ on arc $e$.

	\vspace{-0.2cm}
	The workload of a CNN subgraph is captured with a workload matrix $\boldsymbol{W} \in \mathbb{Z}^{|E|\times|V|}$ with $\boldsymbol{W}(e,v)$ holding the elements to be produced or consumed by node $v$ on arc $e$. A partitioned CNN with $N_{W}$ subgraphs is associated with a workload tuple $W = <\boldsymbol{W}_i ~|~ i \in [1,N_W]>$, with one matrix per subgraph. At each stage, the workload is $f_{\text{in}}N_{\text{out}}K^2h_{\text{out}}w_{\text{out}}$ elements for convolutional and $N_{\text{out}}K^2h_{\text{out}}w_{\text{out}}$ elements for pooling layers with $N_{\text{out}}$ ($h_{\text{out}}$$\times$$w_{\text{out}}$)-sized output feature maps. In the case of $N$ CNNs, the multi-CNN architecture is represented as $G_{\text{multiCE}}$$=$$\{G^1_{\text{CE}}, ..., G^{N}_{\text{CE}}\}$ with multi-CNN topology and workload tuples $\Gamma=<\boldsymbol{\Gamma}_{i} \in \mathbb{R}^{|E_i| \times |V_i|} ~|~ i \in [1,N]>$ 
	and $W = <\boldsymbol{W}_{i,j} \in \mathbb{Z}^{|E_i| \times |V_i|} ~|~ i \in [1,N], j \in [1,N_{W_i}]>$. 
	The initiation interval matrix for the j-th subgraph of the i-th CNN is constructed as $\boldsymbol{II}_{i,j}$$=$$\boldsymbol{W}_{i,j} \oslash \boldsymbol{\Gamma}_i$, and the execution time of a single (j-th) subgraph 
	and all subgraphs 
	of the i-th CNN on the i-th engine are given by Eq. (\ref{eq:single}) and (\ref{eq:all}) respectively: 
	\small
	\begin{align}
	\label{eq:single}
	&t_{i,j}(B, \boldsymbol{\Gamma}_i, \boldsymbol{W}_{i,j}) = \frac{1}{\text{clock\ rate}} \cdot (D_i +  II^{\text{max}}_{i,j} \cdot (B-1)) \\		\label{eq:all}
	&t_{\text{total}}^i(B, \boldsymbol{\Gamma}_i,\boldsymbol{W}_{i,:}) = \sum_{j=1}^{N_{W_i}}t_{i,j}(B,\boldsymbol{\Gamma}_i,\boldsymbol{W}_{i,j})\text{$+$}\sum_{j=1}^{N_{W_i}}t_{i,j,\text{weights}}
	\end{align}
	\normalsize
where $II^{\text{max}}_{i,j}$ is the maximum element of $\boldsymbol{II}_{i,j}$, $B$ the batch size, $D_i$ the pipeline depth of the i-th CNN engine and $t_{i,j,\text{weights}}$ the time to load the weights of the j-th subgraph of the i-th CNN. Moreover, the latency of the j-th subgraph on the i-th engine is given by $L(B$$=$$1,\boldsymbol{\Gamma}_i,\boldsymbol{W}_{i,j})=t_{i,j}(1,\boldsymbol{\Gamma}_i, \boldsymbol{W}_{i,j})$.
	
	
	
	\begin{figure}[t]
		\vspace{-0.22cm}
		\centering
		\includegraphics[trim={0cm 1.5cm 0cm 0cm},clip,width=0.4\textwidth]{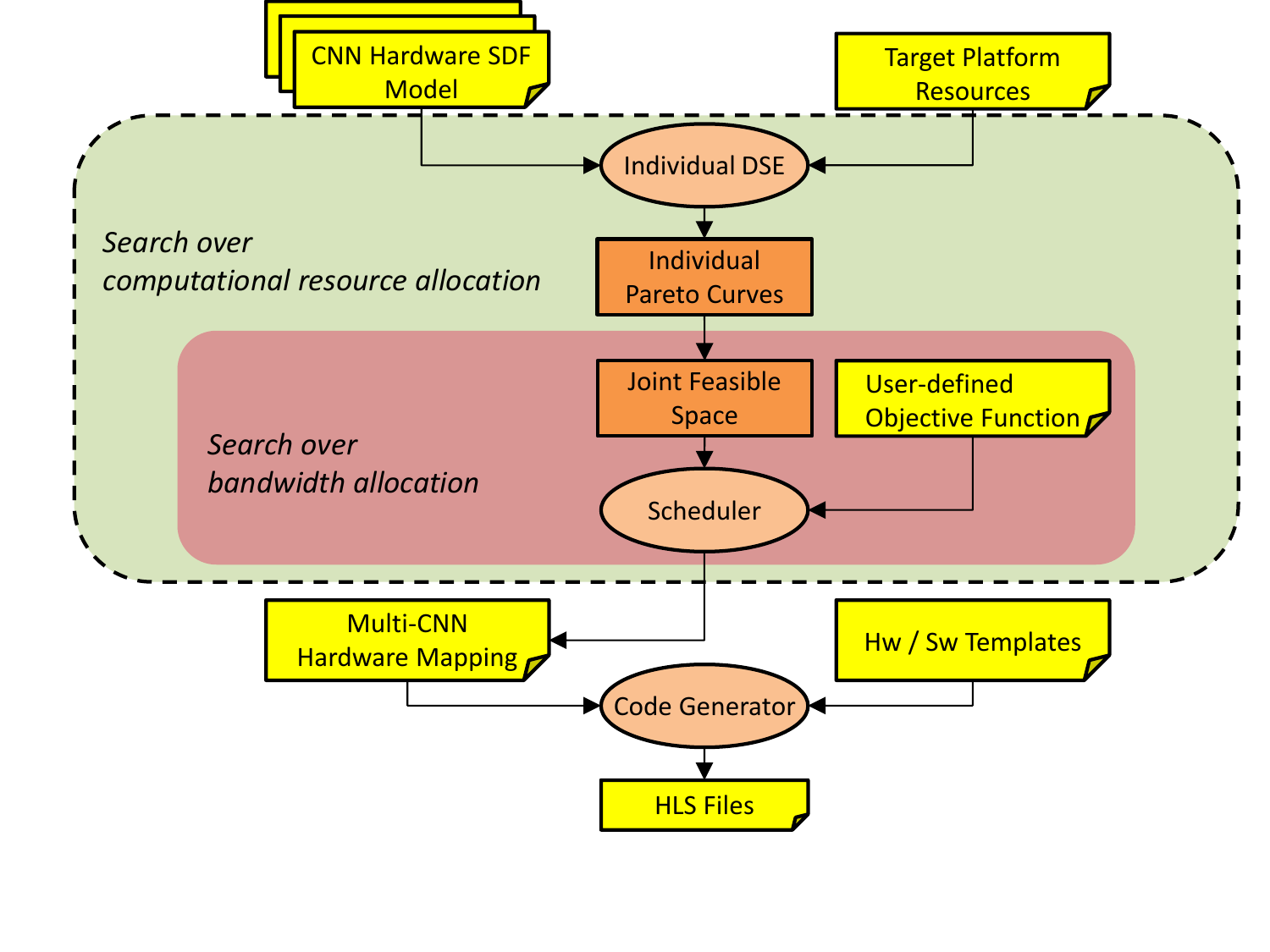}
		\vspace{-0.3cm}
		\caption{Overview of f-CN$\text{N}^{\text{x}}$'s DSE flow.}
		\label{dse_flow}
		\vspace{-0.9cm}
	\end{figure}
	
	\vspace{-0.2cm}
	\textbf{Search Method.} Fig. \ref{dse_flow} shows the proposed DSE method. First, by exploring the design space of each individual CNN on the resource budget of the target FPGA, the design points on the latency-resource Pareto front of each CNN are found, without accounting for the shared bandwidth {\color{black}to the external memory}. 
	Each individual design point corresponds to different 1) partitioning of the CNN, 2) structure of the pipeline and \mbox{3) folding} factors for each hardware stage, and is characterised by its performance, 
	on-chip resource consumption and its workload, including the computational and off-chip memory bandwidth requirements of its subgraphs. 
	
	\vspace{-0.2cm}
	As a next step, f-CNN$^\text{x}$ performs an enumeration of all the combinations of design points that belong to the Pareto fronts of individual CNNs to obtain \textit{joint} design points, denoted by $\sigma$. The combinations that do not lie in the \textit{feasible space} of the target FPGA are discarded based on their aggregate on-chip resource consumption as $\sum_{i=1}^{N}\boldsymbol{rsc}(\sigma_i) \le \boldsymbol{rsc}_{\text{Avail.}}$, 
	where $\sigma_i$ denotes the hardware design for the i-th CNN, $N$ the number of CNNs and $\boldsymbol{rsc}(\sigma_i)$ the resource consumption vector, including LUTs, Flip-Flops, DSPs and BRAMs. Next, the scheduler module (Fig. \ref{dse_flow}) takes into account the sharing of the bandwidth and traverses the feasible space to search for the (joint design point, memory transfers schedule) pair that optimises a user-defined objective function. 
	After the highest performing joint design point has been selected, the corresponding multi-CNN architecture is implemented using an automated code generation mechanism.
	
	\vspace{-0.3cm}
	\subsection{Scheduler}
	\vspace{-0.1cm}
	\label{sec:sched}
	The scheduler is responsible for taking into account the effect of the shared memory bandwidth and identifying the highest performing design for the multi-CNN architecture based on a user-defined objective function. This module takes as input the joint design points of the Pareto front and predicts the actual performance of each point after scheduling the memory transfers. 
	In this respect, the quality of the memory access schedule affects substantially the utilisation of the architecture, {\color{black}especially} in cases with high bandwidth contention. 
	
	\vspace{-0.2cm}
	To this end, we cast the time-sharing of the external memory bandwidth as a \textit{cyclic scheduling} problem \cite{Draper1999} due to the constant stream of new inputs to the CNNs. Based on this formulation, a set of tasks, in this case CNN inferences, have to be performed repeatedly. The solution of the cyclic scheduling problem would yield a schedule for all tasks in the presence of precedence and resource sharing constraints. In our formulation, the precedence constraints are the dependencies between the subgraphs of each CNN with resource sharing focusing on the off-chip memory bandwidth. Moreover, we require our solution to be periodic with a fixed period, named \textit{cycle time}, and hence allow each CNN to repeat multiple times during one cycle time.
	Formally, we pose the following cyclic scheduling problem.
	\vspace{-0.3cm}\\
	\\
	\noindent
	\textbf{Inputs:}
	\vspace{-0.2cm}
	\begin{itemize}
		\itemsep0.2em 
		\item $N$: the number of CNNs,
		\item $N_{W_{i}}, i \in [1,N]$: the number of subgraphs of each CNN,  
		\item $S = \{s_{i,j} ~|~ i \in [1,N], j \in [1,N_{W_i}]  ~\}$: the set of subgraphs,
		\item $L(s)$: the latency of each subgraph,
		\item $b(s)$: the memory bandwidth usage for each subgraph,
		\item ${s_{i,j} < s_{i,j+1}, ...}$ : the set of precedence constraints on subgraphs,
		\item $K$: the cycle time (or schedule period),
		\item $rep(i), ~ i \in [1,N]$: the repetitions of each CNN inference in a cycle time,
		\item $B_{\text{mem}}$: the available memory bandwidth.
	\end{itemize}
	\vspace{-0.1cm}
	By allowing multiple repetitions of each CNN within a cycle time, the augmented set of subgraphs becomes:
	\vspace{-0.1cm}
	\begin{equation*}
	S_{\text{aug}} = \{s_{i,j} ~|~ i \in [1,N], j \in [1,rep(i)N_{W_i}]\}
	\end{equation*}
	\noindent
	\textbf{Decision variables:}
	\vspace{-0.2cm}
	\begin{itemize}
		\item $st(s) \in [0,K)$, $s \in S_{\text{aug}}$: start time of each subgraph.
	\end{itemize}
	\vspace{-0.1cm}	
	\noindent
	In addition, we define the following constraints:
	\vspace{-0.1cm}
	\begin{enumerate}
		\itemsep0.05em 
		\item All subgraphs must be scheduled and the start time of each subgraph must lie within the cycle time:
		\vspace{-0.1cm}
		\begin{equation*}
		0 \le st(s) < K, \; s \in S_{\text{aug}}
		\end{equation*}
		\item If 
		subgraph $s_i$ precedes 
		$s_j$, then start time of $s_j$ must occur after the end time of $s_i$ within the cycle time:
		\vspace{-0.15cm}
		\begin{equation*}
		s_i < s_j \Rightarrow st(s_i) + L(s_i) < st(s_j)
		\end{equation*}
		\vspace{-0.5cm}
		\item The memory bandwidth utilisation of subgraphs that are scheduled during the same slot must not exceed the available bandwidth, to minimise contention. 
		
	\end{enumerate}

	
	\vspace{-0.2cm}
	\textbf{Slow-down Scheduler.}
{As described in Sec. \ref{sec:opps_section}, due to the structure of CNNs, the scheduling of memory transfers offers an \mbox{opportunity} for optimisation. Although the on-chip 
resources constitute a \textit{hard} constraint which cannot be violated by the aggregate consumption of the CNN engines
, memory bandwidth is a \textit{soft} constraint and can be violated from a design by requiring more bandwidth than is available. Nevertheless, bandwidth violations lead to memory contention between the CNN engines, and therefore, if allowed, the estimated performance from the performance model would be different to the actual measured performance, making the DSE irrelevant. Additionally, if we impose bandwidth as a hard constraint and schedule the subgraphs to ensure no violations, the bandwidth will be underutilised, due to the conservative scheduling and the discrete nature of the subgraphs. To alleviate this, we introduce a control mechanism over the processing rate of each CNN engine at any time instant, which is optimised to remove memory violations while maximising bandwidth utilisation. 
	}
	\vspace{-0.1cm}
	
	Classic scheduling algorithms, such as Integer Linear Programming (ILP) 
	and heuristic schedulers, treat each schedulable unit in a faithful manner, without modifying its execution time and bandwidth requirements. Due to this property, such schedulers do not exhibit the flexibility and expressive power that can exploit the per-cycle deterministic control offered by FPGAs over memory transfers. 
	To this end, we propose a rate-controlling scheduler which 
	controls the processing rate of each CNN engine at any instant. We model this by introducing an additional set of decision variables to our cyclic scheduling problem, under the name \textit{slow-downs}, defined as in Eq. (\ref{slow_down_equ}). 
	\vspace{-0.1cm}
	\begin{equation}
	\label{slow_down_equ}
	sl_{i,j} \in (0,1], \; i \in [1,N], j \in [1, rep(i)N_{W_{i}}]
	\end{equation}
	\vspace{-0.4cm}
	\begin{equation}
	\label{slow_down_latency_bw_equ}
	\begin{cases}
	L'(s_{i,j}) = \frac{1}{sl_{i,j}} \times L(s_{i,j})\\
	b'(s_{i,j}) = sl_{i,j} \times b(s_{i,j})
	\end{cases}
	\end{equation}
	We interpret slow-downs as a control factor over the bandwidth allocated to each CNN engine at each time instant. With the pipelines of our architecture operating under a data-driven paradigm, a slower input data rate would slow down the processing speed of an engine and, at the same time, reduce the bandwidth requirements imposed on the off-chip memory by a particular subgraph (Eq. (\ref{slow_down_latency_bw_equ})). As a result, with this formulation, a subgraph with bandwidth violations can be slowed down and potentially be scheduled more efficiently to better reflect the actual attainable performance upon deployment.

\begin{figure}[t]
		\vspace{-0.3cm}
		\centering
		\includegraphics[trim={0cm 3.5cm 14cm 3cm},clip,width=0.475\textwidth]{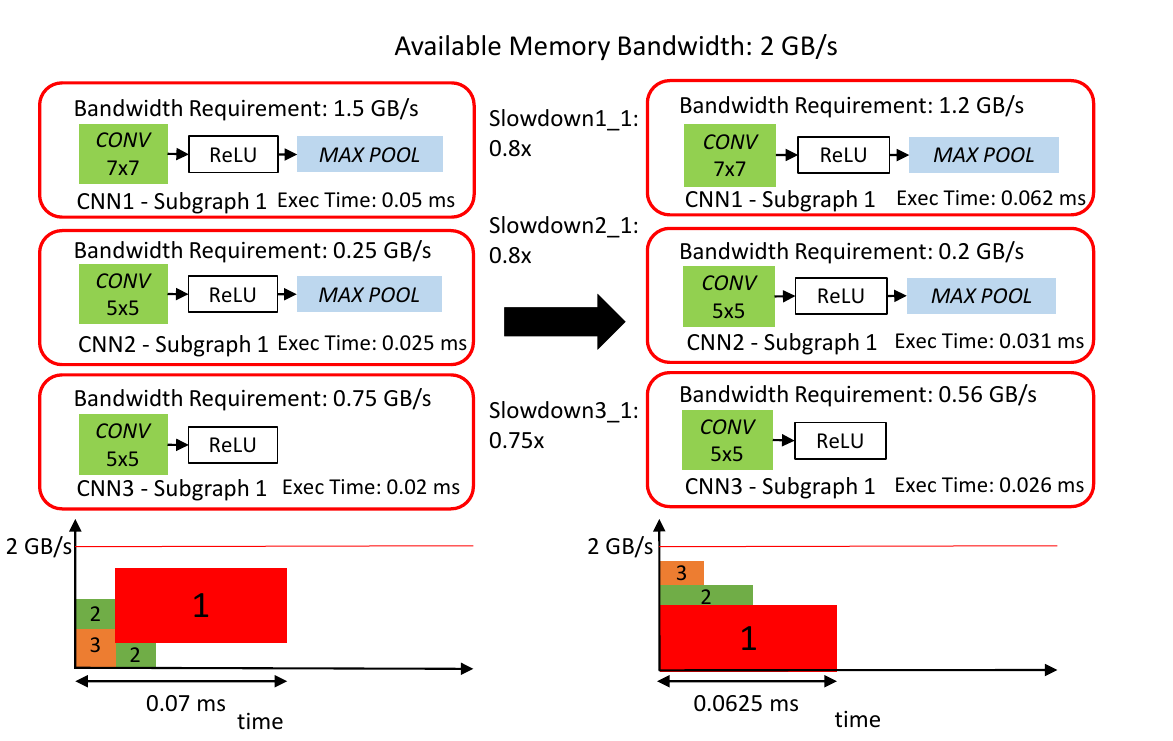}
		\vspace{-0.3cm}
		\caption{An example of the effect of the proposed slow-downs.}
		\label{slow-down_diagram}
		\vspace{-0.85cm}
	\end{figure}

	\vspace{-0.15cm}
	Fig. \ref{slow-down_diagram} illustrates the potential benefits of slow-downs in the case of three CNNs. {The bottom left image shows the predicted performance if no slow-downs were introduced and no bandwidth violations were allowed.} In this scenario, the aggregate required bandwidth of the three subgraphs exceeds the available budget by 1.25$\times$ and the subgraphs cannot be scheduled in parallel without causing contention, leading to the schedule depicted on the bottom left of Fig. \ref{slow-down_diagram}. By applying slow-down factors of 0.8, 0.8 and 0.75 respectively, 80\% of the required bandwidth is supplied to the first two subgraphs and 75\% to the third and, in this way, the processing rate of each CNN engine is decreased proportionally. 
	This approach decreases the aggregate required bandwidth to the feasible \mbox{1.96 GB/s}, 
	leading to a shorter schedule.
	
	\vspace{-0.15cm}
	The extension of the multi-CNN cyclic scheduling problem to include slow-downs expands further the number of design parameters that we have to optimise, leading to a more complex design space. To solve the scheduling problem, we treat it as multiobjective optimisation (MOO) with an objective function that assesses the quality of a joint design point after scheduling. The objective function is user-defined and can be selected to capture the application-level importance of each CNN. Two characteristic objective functions are shown in \mbox{Eq. (\ref{eq:fpsobj})} and (\ref{eq:maxthrpt}).
	\begin{myobj}
		\label{loco}
		\textbf{FPSobj}: Optimise the multi-CNN mapping to achieve the target frame rate in frames per second (fps) for each CNN, with equal importance across the CNNs.
		\vspace{-0.2cm}
		\begin{align}
		\label{eq:fpsobj}
		&\min_{\{\sigma_i\}_{1 \le i \le N}} \sum_{i=1}^{N}\left(\frac{fps(\sigma_i)-fps^{\text{target}}_i}{fps^{\text{target}}_i}\right)^2 \\ 
		&\text{ s.t. } \sum_{i=1}^{N}\boldsymbol{rsc}(\sigma_i) \le \boldsymbol{rsc}_{\text{Avail.}} \nonumber
		\vspace{-2cm}
		\end{align}
		\vspace{-0.65cm}
	\end{myobj}
	\noindent
	where $fps(\sigma_i)$ is the fps of the $\sigma_i$ design point of the i-th CNN given the shared bandwidth constraints, $fps^{\text{target}}_i$ is set to $\min(fps^{\text{user}}_i,fps^{\text{max}}_i)$, i.e the minimum between the user-defined target fps and the maximum attainable fps\footnote{$fps^{\text{max}}_i$ is the maximum attainable fps for the i-th network assuming the whole device and off-chip memory bandwidth are available only to this CNN.} for the i-th network on the target platform. The fps of each design point $\sigma_i$ is divided by the target fps to obtain a non-dimensional objective function and place equal weight to all the CNNs.
	
	\begin{myobj}
		\textbf{MaxThrpt}: Optimise the multi-CNN mapping to achieve the maximum throughput in GOp/s for each CNN that lies in the joint design space.
		\vspace{-0.2cm}
		\begin{align}
	 	\label{eq:maxthrpt}
	 	&\min_{\{\sigma_i\}_{1 \le i \le N}} \sum_{i=1}^{N}\left(\frac{T(\sigma_i)-T^{\text{max}}_i}{T^{\text{max}}_i}\right)^2 \\
	 	&\text{ s.t. } \sum_{i=1}^{N}\boldsymbol{rsc}(\sigma_i) \le \boldsymbol{rsc}_{\text{Avail.}} \nonumber
	 	\vspace{-2cm}
	 	\end{align}
	 	\vspace{-0.65cm}
	\end{myobj}
	\noindent
	where $T(\sigma_i)$ denotes the throughput of the $\sigma_i$ design point of the i-th CNN in GOp/s given the shared bandwidth constraints and $T^{\text{max}}_i$ the maximum attainable throughput for the i-th CNN on the target FPGA. 
	The throughput of each $\sigma_i$ is divided by the maximum throughput to obtain a non-dimensional objective function and place equal weight to all the CNNs. 
	
%
\vspace{-0.2cm}
	The resource-constrained cyclic scheduling problem has been proven to be NP-hard \cite{LEVNER2010}. In our multiple CNN formulation of Sec. \ref{sec:sched}, which is used to obtain a schedule for each multi-CNN design point, the size of the problem is proportional to the number of subgraphs to be scheduled. For small-sized problems, we model the problem as an integer linear program (ILP) and employ an ILP solver to obtain the optimal solution. The excessive runtime of ILP solvers sets a limit on the scale of solvable problems and, therefore, in such cases, a heuristic scheduler is required to obtain a solution. To this end, we developed a heuristic scheduler that combines Resource Constrained List Scheduling (RCLS) \cite{Micheli1994} with slow-downs. {\color{black} With this approach, given a joint design point and a set of slow-downs, the lowest latency schedule is obtained.
	 	
	 \vspace{-0.2cm}
	 \textbf{Memory-aware DSE.} To select the highest performing schedule for each point, we developed an iterative, derivative-free \textit{pattern search} (PS) optimiser \cite{Audet_2002} that, given a joint design point $\sigma$, initial slow-down vector $\boldsymbol{sl}_0$, memory bandwidth budget $B_{\text{mem}}$ and target objective function $F$, searches over slow-downs. At each 2-step iteration, the optimiser first \textit{explores} neighbouring solutions of the slow-down vector $\boldsymbol{sl}$ in a finite number of directions. If a solution that improves $F$ is found, the optimiser updates the slow-down values. Else, a \textit{polling} step is performed to search for candidate solutions farther away from the current $\boldsymbol{sl}$. The PS algorithm requires a large number of direct evaluations of $F$, which are efficiently performed by means of the slow-down scheduler and the SDF performance model (Sec. \ref{sec:dse}). In this manner, the highest performing schedule in terms of $\boldsymbol{sl}$ is obtained for each $\sigma$}. 

		\setlength{\textfloatsep}{0pt}
		\begin{algorithm}[!t]	
			\footnotesize
			\SetAlgoLined
			\LinesNumbered
			\DontPrintSemicolon
			\KwIn{Set of joint design points $\Sigma$ in the feasible space}
			\nonl
			\myinput{Objective function $F(\sigma, \boldsymbol{sl})$, $\sigma \in \Sigma$}
			\nonl
			\myinput{Off-chip memory bandwidth budget $B_{\text{mem}}$}
			\KwOut{Joint design point $\sigma^*$ chosen for the architecture} 
			\nonl
			\myoutput{Optimised slow-down factors $\boldsymbol{sl}^*$ for $\sigma^*$}
			\ForEach{joint design point $\sigma \in \Sigma$}{
				\texttt{/*} - - - \textit{slow-down initialisation proposals} - - - \texttt{*/}\;
				$sched_{\text{init}} \leftarrow \text{RCLS}(\sigma);$ // \textit{Without bandwidth constraints}\; 
				$viol(s) \leftarrow \text{Violations}(\sigma, sched_{\text{init}},B_{\text{mem}}), \;\forall s \in \sigma$ \;
				$\boldsymbol{sl}_0(s) \leftarrow \text{RemoveViolations}(s,viol(s)), \; \forall s \in \sigma$\;
				\nonl
				\;
				\texttt{/*} - - - \textit{slow-down search} - - - \texttt{*/}\;
				Apply a pattern search algorithm over the slow-downs to \;
				\nonl 
				to optimise for $F$: \;
				$[\boldsymbol{sl}, F(\sigma,\boldsymbol{sl})] \leftarrow \text{PatternSearch}(\sigma, \boldsymbol{sl}_0, B_{\text{mem}}, F)$\;
				\If{$F$ improved}{
					$\sigma^* \leftarrow \sigma; \quad \boldsymbol{sl}^* \leftarrow \boldsymbol{sl}$\;
				}
			} 
			\caption{\small Memory-aware DSE for multiple CNNs}
			\label{dse_algo}
		\end{algorithm}
	\afterpage{\global\setlength{\textfloatsep}{\oldtextfloatsep}}

	\vspace{-0.2cm}
	{Algorithm \ref{dse_algo} presents the overall memory-aware DSE, searching over both on-chip resource and external memory bandwidth allocations. The DSE searches over different on-chip resource allocations between CNN engines (line 1). For each allocation, the highest performing schedule is found by means of the PS optimiser (lines 7-8). 
	Prior to the optimiser, a greedy strategy is employed to generate \textit{slow-down proposals} (lines 3-5) that place $\boldsymbol{sl}_0$ in a region of the design space with no violations, in order to facilitate the slow-down search. At the end of the loop, the (architecture, schedule) pair that optimises $F$ is selected. Further details of the slow-down scheduler and the PS optimiser are omitted due to space constraints.}
	
\begin{figure}[t]
		\vspace{-0.6cm}
		\centering
		\includegraphics[trim={1.9cm 4cm 6cm 2.1cm},clip,width=0.4\textwidth]{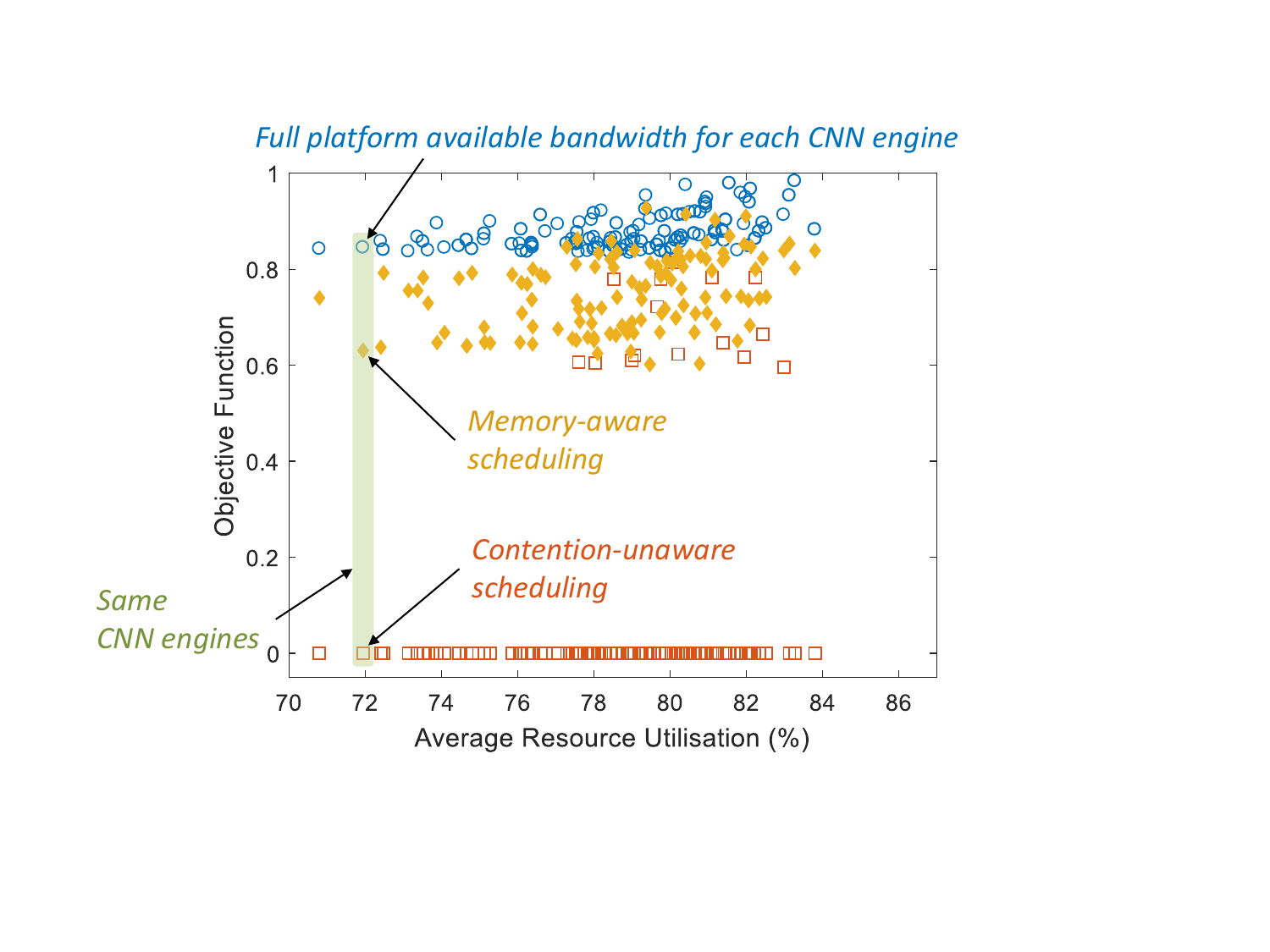}
		\vspace{-0.1cm}
		\caption{ 
		Effect of the proposed DSE (Table \ref{table:comp_base}, benchmark 7). 
		}
		\label{fig:dse_change}
		\vspace{-0.3cm}
	\end{figure}

	\vspace{-0.2cm}
	{\color{black}To illustrate the impact of the proposed memory-aware scheme, \mbox{Fig. \ref{fig:dse_change}} depicts how the memory-aware design shifts the candidate joint design points to regions with improved objective function values for benchmark 7 of \mbox{Table \ref{table:comp_base}}. The horizontal axis shows the average resource usage across LUTs, FFs, DSPs and BRAMs on Zynq XC7Z045. 
	The explored joint design points appear in (blue, red, yellow) triplets. The points of a triplet have the same on-chip resource allocation, but different scheduling. Blue points correspond to the peak performance if each CNN engine had access to the full platform bandwidth. Red points show the case when each engine attempts to access the external memory asynchronously. 
	In contrast to the contention-unaware red points, the memory-aware design enables yellow points to tailor the memory access policy to the target multiobjective criterion and match it to the performance requirements of each CNN, and as a result outperform red points. 
	}


	\vspace{-0.2cm}
	\subsection{Multi-CNN Hardware Scheduler}
	\vspace{-0.1cm}
	\label{sec:mcnnhs}
 	The selected schedule is mapped to hardware with a rate-controlling mechanism and a multi-CNN hardware scheduler.
	
	\vspace{-0.4cm}
	\textbf{Rate-controlling Mechanism.}
	To implement a (schedule, slow-downs) pair, each CNN engine has to be supplied a specific fraction of the available bandwidth at each time instant. To this end, we discretise time into slots of equal size. During a slot, only a single CNN engine is allocated the available bandwidth, with all engines served in a round-robin fashion. By allowing the CNN engines to occupy several consecutive slots, a tunable fraction of the bandwidth is provided to each engine during each period of slots as given by Eq. (\ref{frac_bw}).\\
	\vspace{-0.1cm}
	\begin{small}
	\begin{equation}
	\label{frac_bw}
	B(s_{i,j}) = \frac{slots(s_{i,j})}{\#\ slotsTotal}B_{\text{mem}}, \; i \in [1,N], j \in [1, N_{W_i}]
	\vspace{-0.2cm}
	\end{equation}
	\end{small}
	where $B(s_{i,j})$ is the average supplied bandwidth and $slots(s_{i,j})$ is the number of consecutive slots assigned during the execution of the j-th subgraph by the i-th CNN engine. With this formulation, to comply with a selected (schedule, slow-downs) pair, the supplied bandwidth $B(s_{i,j})$ has to be equal to the required bandwidth $b'(s_{i,j})$ (Eq. (\ref{slow_down_latency_bw_equ})). Hence, the values of $slots(s_{i,j})$ are found by solving Eq. (\ref{frac_bw}) with $B(s_{i,j})$ set equal to $b'(s_{i,j})$. Finally, the size of each slot in cycles is equal to the selected burst length for the memory transfers and is discussed in the following section.
	
	
%

\begin{figure}[t]
		\vspace{-0.6cm}
		\centering
		\includegraphics[trim={0cm 1.8cm 0cm 3cm},clip,width=0.48\textwidth]{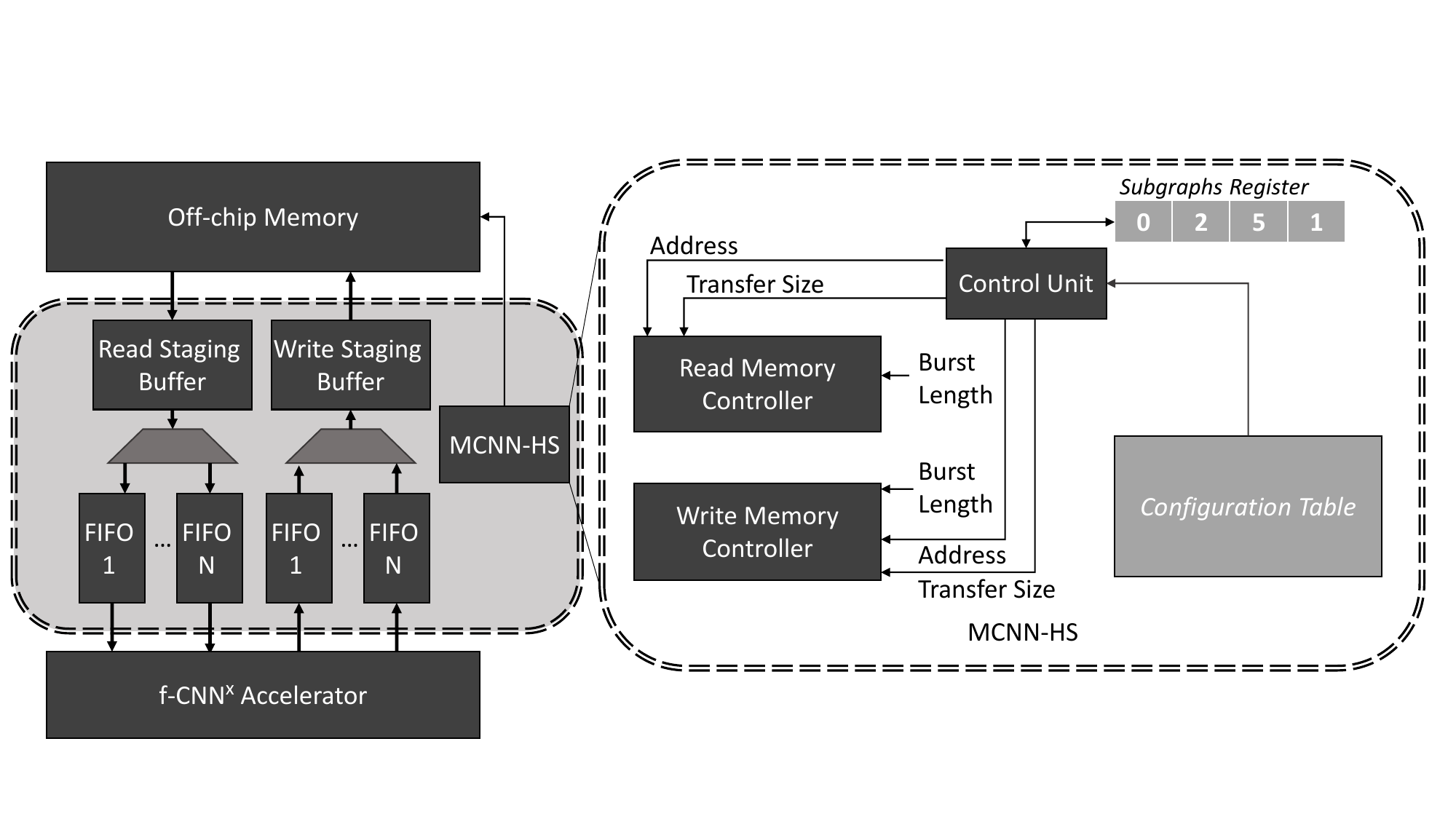}
		\vspace{-0.5cm}
		\caption{\mbox{Microarchitecture of the multi-CNN hardware scheduler.}}
		\label{mem_schdl_fig}
		\vspace{-0.2cm}
	\end{figure}

	\vspace{-0.2cm}
	\textbf{Microarchitecture.}
	{\color{black}Key enabler of the proposed design is the \texttt{MCNN-HS} module that is responsible for interfacing the CNN engines with the external memory.} Fig. \ref{mem_schdl_fig} shows the microarchitecture of \texttt{MCNN-HS}. 
%
	The selected schedule is encoded into a compile-time configuration of \texttt{MCNN-HS} by means of the rate-controlling mechanism. The \texttt{MCNN-HS} communicates with the external memory via two memory controllers and hosts two staging buffers that mediate between the external memory and the FIFOs of the CNN engines. The sizes of the staging buffers are determined based on the largest on-chip storage requirement among the target subgraphs. {Moreover, the FIFOs are employed to smooth out the time discretisation of the external memory accesses, so that the CNN engines see a continuous flow of data, instead of bursts, with their depth configured based on the processing rate of each engine.}

	\vspace{-0.2cm}
	\texttt{MCNN-HS} comprises a configuration table and a control unit (CU). The configuration table stores encoded information for each subgraph about the amount of data to be transferred, the allocated number of consecutive slots and the off-chip memory addresses, with the contents of the table determined at compile time by the rate-controlling mechanism. The CU is responsible for orchestrating the multi-CNN schedule at run time. A subgraphs register is used to keep track of the currently active subgraph for each CNN and look up the appropriate entries of the configuration table. While all the CNN engines operate in parallel, {\color{black}off-chip memory} access is supplied to each engine in a round-robin manner by the CU. The burst lengths, and hence the duration of a slot in cycles, for the memory controllers are set to a fixed value across all transactions in order to simplify their configuration and minimise memory access inefficiencies\footnote{By investigating the impact of burst length on bandwidth utilisation efficiency, a burst length of 1024 was selected for \texttt{MCNN-HS}, achieving higher than 90\% measured efficiency on ZC706.}. Finally, if the wordlength 
	is smaller that the width of the memory port, multiple values are packed together to increase bandwidth utilisation.

	\vspace{-0.2cm}
	As an example of \texttt{MCNN-HS}'s operation, consider a setting with three CNNs with one subgraph each, $slots(s_{1,1})$$=$$1$, $slots(s_{2,1})$$=$$2$ and $slots(s_{3,1})$$=$$4$, $16384$, $16384$ and $32768$ elements to be read, a burst length of $1024$, 16-bit 
	precision and a shared 64-bit memory port. With 16-bit values packed in groups of 4, the 64-bit port 
	transfers 
	$4$ elements per cycle. Given the burst length of $1024$, subgraphs $s_{1,1}$, $s_{2,1}$ and $s_{3,1}$ are supplied $1024$, $2048$ and $4096$ consecutive cycles respectively with a transfer rate of $4$ elements/cycle. 
	Overall, in a period of $7$ slots, the subgraphs will receive 4096, 8192 and 16384 elements in a round-robin fashion. To receive all their data, all slots have to execute $4$, $2$ and $2$ times respectively. The fraction of supplied bandwidth in this case would be 14.28\%, 28.57\% and 57.14\% respectively.
	\vspace{-0.5cm}
	



	

	\begin{table}[h]
	\centering
	
	\caption{Benchmarks}
	\vspace{-0.25cm}
	\label{benchmarks_table}
	\resizebox{0.8\linewidth}{!}{%
		\setlength\tabcolsep{2pt} 
		\begin{tabular}{l l c r l}
			\toprule
			\multirow{1}{*}{Model Name} & &
			\multicolumn{1}{c}{Layers} & \multicolumn{1}{c}{\multirow{1}{*}{Workload}} & \multicolumn{1}{l}{ \multirow{1}{*}{Task}}                                   \\ 
			\midrule 
			
			LeNet-5 (Caffe version) 
			& \cite{Lecun1998}                   & 4                                            & 0.0038 GOps                       
			&  \multicolumn{1}{l}{\begin{tabular}[c]{@{}l@{}}  Digit Recognition\end{tabular}}   \\ 
			
			CIFAR-10 & \cite{Krizhevsky09learningmultiple}     & 9                                            & 0.0247 GOps  
			& Object Recognition             \\ 
			PilotNet & \cite{nvidia_2016} & 10 & 0.0620 GOps & Wheel Stirring \\
			ZFNet & \cite{Zeiler_2014} & 10 & 2.2219 GOps & Object Detection \\
			SceneLabelCNN & \cite{Cavigelli2015}        & 8                                            & 7.6528 GOps      
			& Scene Labelling                \\ 
			VGG16 & \cite{Simonyan14c} 
			& 31 & 30.7200 GOps 
			& Scene Recognition \\
			
			\bottomrule
		\end{tabular}%
	}
	\vspace{-0.2cm}
\end{table}

\setcounter{table}{2}

\begin{table*}[t]
	\vspace{-0.2cm}
	\centering
	\caption{Comparison of f-CN$\text{N}^{\text{x}}$ and contention-unaware FPGA accelerator (batch size = 1)}
	\vspace{-0.15cm}
	\label{table:comp_base}
	\footnotesize
	\resizebox{0.8\linewidth}{!}{
		\begin{tabular}{c c l l l l c c}
			\toprule
			ID & Benchmark & \multicolumn{1}{l}{Model Set} & \begin{tabular}[c]{@{}l@{}} Available \\ Bandwidth \end{tabular} &  \multicolumn{1}{l}{Baseline (GOp/s)} & f-CN$\text{N}^{\text{x}}$ (GOp/s) & \begin{tabular}[c]{@{}l@{}} Speed-up \\ (geo. mean) \end{tabular} &\begin{tabular}[c]{@{}l@{}} \textbf{FPSobj} \\ (\% Gain) \end{tabular} \\
			\midrule
			1 & 3 CNNs & ZFNet, SceneLabelCNN, VGG16 & 1.0 GB/s & (15.43, 28.61, 16.40) & (13.97, 60.14, 48.27) & 77\% & 42\% \\
			2 & 3 CNNs & ZFNet, SceneLabelCNN, VGG16 & 1.7 GB/s & (17.03, 91.23, 26.15) & (19.92, 85.71, 68.80) & 42\% & 51\% \\
			3 & 3 CNNs & ZFNet, SceneLabelCNN, VGG16 & 2.0 GB/s & (22.58, 87.48, 39.01) & (21.45, 92.30, 74.08) & 24\% & 38\% \\
			4 & 3 CNNs & ZFNet, SceneLabelCNN, VGG16 & 3.8 GB/s & (22.70, 96.22, 48.76) & (23.05, 99.21, 79.63) & 19\% & 37\% \\
			%
			%
			
			\midrule
			5 & 4 CNNs & ZFNet, PilotNet, SceneLabelCNN, VGG16   & 1.0 GB/s & (\phantom{0}8.12, \phantom{0}0.72, 33.58, 11.22) & (10.39, \phantom{0}1.26, 47.71, 47.87) & 91\% & 54\% \\
			6 & 4 CNNs & ZFNet, PilotNet, SceneLabelCNN, VGG16   & 1.7 GB/s & (13.51, \phantom{0}1.27, 58.14, 23.33) & (21.18, \phantom{0}1.87, 72.91, 48.77) & 57\% & 43\% \\
			7 & 4 CNNs & ZFNet, PilotNet, SceneLabelCNN, VGG16   & 2.0 GB/s & (16.00, \phantom{0}1.47, 68.11, 30.37) & (20.00, \phantom{0}1.95, 68.86, 69.08) & 40\% & 40\% \\
			8 & 4 CNNs & ZFNet, PilotNet, SceneLabelCNN, VGG16   & 3.8 GB/s & (15.46, \phantom{0}1.61, 85.14, 37.96) & (16.28, \phantom{0}1.97, 93.43, 75.00) & 29\% & 32\% \\
			\bottomrule
		\end{tabular}
	}
	\vspace{-0.5cm}
\end{table*}

	\vspace{-0.25cm}
	
	\section{Evaluation}
	\label{evaluation_sec}
	\vspace{-0.2cm}
	\subsection{Experimental Setup}
	\vspace{-0.2cm}
	In our experiments, we target the 
	ZC706 board mounting the Zynq XC7Z045 SoC, with a clock rate of 150 MHz. All hardware designs were synthesised and placed-and-routed with Xilinx's Vivado Design Suite (v17.2) and run on the ZC706 board. The ARM CPU was used to measure the performance of each design. For the evaluation, Q8.8 16-bit 
	precision was used which has been studied to give similar results to 32-bit floating-point \cite{Venieris_2017b}. {In each multi-CNN benchmark (\mbox{Tables \ref{rcls_vs_ilp_table} and \ref{table:comp_base}}), the available bandwidth was controlled by using a different number of memory ports and amount of word packing.}

	\vspace{-0.2cm}
	\mbox{Table \ref{benchmarks_table}} lists our benchmark CNNs. 
	\mbox{LeNet-5} and \mbox{CIFAR-10} have 
	comparatively small workloads and are employed to evaluate the RCLS against the optimal ILP scheduler. 
	PilotNet, ZFNet, SceneLabelCNN and VGG16 
	pose mapping challenges such as the non-uniform filters of ZFNet, the large filters of SceneLabelCNN and the computational intensity of VGG16. Moreover, ZFNet and VGG16 are used for numerous object detectors \cite{Ren_2017} in multi-CNN applications, with VGG16's pretrained model widely employed in new domains \cite{Badrinarayanan_2017}.
	
	\vspace{-0.2cm}
	The rest of this section focuses on 1) the evaluation of the proposed heuristic scheduler with respect to an optimal ILP scheduler, 2) comparisons with a contention-unaware multi-CNN FPGA design and 3) with highly optimised designs targeting an embedded GPU across multi-CNN settings.
	
	
	%
	
	\setcounter{table}{1}
	
	\begin{table}[t]
	\vspace{-0.2cm}
	\centering
	\caption{Proposed vs. Optimal ILP Scheduler}
	\vspace{-0.25cm}
	\label{rcls_vs_ilp_table}
	\footnotesize
	\resizebox{1\linewidth}{!}{
		\setlength\tabcolsep{1pt} 
		\begin{tabular}{c c l c l l l l}
			\toprule
			ID & Benchmark  & \multicolumn{1}{l}{Model Set} & Subgraphs & \begin{tabular}[c]{@{}l@{}} Available \\ Bandwidth \end{tabular} & \begin{tabular}[c]{@{}c@{}} ILP \\ MaxThrpt/Runtime \end{tabular} 
			& \begin{tabular}[c]{@{}c@{}} RCLS \\ MaxThrpt/Runtime \end{tabular} 
			\\
			\midrule
			1 & 2 CNNs & LeNet-5, CIFAR-10           & 20 & 0.5 GB/s & 0.395 / 5.8 min & 0.395 / 3.6 s \\
			2 & 2 CNNs & LeNet-5, CIFAR-10   		& 18 & 3.8 GB/s & 0.254 / 5.9 min & 0.254 / 3.6 s  \\
			3 & 3 CNNs & LeNet-5, 2$\times$ CIFAR-10 & 44 & 1.5 GB/s & 0.983 / 2h36 & 0.983 / 4.1 min \\
			4 & 3 CNNs & LeNet-5, 2$\times$ CIFAR-10 & 44 & 3.8 GB/s & 0.946 / 2h30 & 0.946 / 2.5 min \\
			5 & 4 CNNs & LeNet-5, 3$\times$ CIFAR-10 & 548 & 1.5 GB/s & - & 1.875 / 1h \\
			6 & 4 CNNs & LeNet-5, 3$\times$ CIFAR-10 & 1454 & 3.8 GB/s & - & 1.829 / 1h &              \\
			\bottomrule
		\end{tabular}
	}
\end{table}

\setcounter{table}{3}
	
	\vspace{-0.35cm}
	\subsection{Evaluation of Proposed Scheduler}
	\vspace{-0.15cm}
	In this section, the quality of the proposed RCLS-based scheduler is assessed. This is investigated by using the \mbox{\textbf{MaxThrpt} criterion} (Eq. (\ref{eq:maxthrpt})) to generate multi-CNN hardware designs using both the RCLS and the ILP schedulers and measuring the real achieved value on the target FPGA board. The comparisons are performed on small-scale problems in order for the ILP solver to yield a solution in a tractable amount of time, where the scale is defined as the number of subgraphs to be scheduled. We employ the low-end \mbox{LeNet-5} and CIFAR-10 and compare across six settings by varying the number of CNNs and the available bandwidth. \mbox{Table \ref{rcls_vs_ilp_table}} presents the measured results on ZC706. 
	The selected multi-CNN designs were implemented and run on the target platform and the measured performances were used to yield the achieved \mbox{\textbf{MaxThrpt}}. {\color{black}The results demonstrate that} both schedulers achieve identical values with respect to the objective function, with the RCLS scheduler generating the schedule in much shorter runtime. When scaling to four CNNs (rows 5 and 6), the problem size 
	increases substantially and the excessive runtime of the ILP solver prohibits us from obtaining an optimal schedule, which verifies the necessity of the heuristic scheduler.

	\vspace{-0.4cm}
	\subsection{Comparison with Contention-unaware FPGA Architecture}
	\vspace{-0.15cm}
	{\color{black}As this is the first work that addresses the problem of mapping multiple CNNs on an FPGA, we cosidered as a baseline the application of the proposed methodology without scheduling optimisation, to yield a contention-unaware implementation. 
	In this respect, we compare the achieved performance of a) the contention-unaware design and b) the f-CNN$^{\text{x}}$ design generated using the complete methodology, on a number of multi-CNN benchmarks.}
	The contention-unaware design comprises the highest performing f-CNN$^{\text{x}}$ architecture with the CNN engines configured so that their aggregate on-chip resource consumption is feasible on the target FPGA, but without exposing the sharing of the bandwidth to the DSE. 
	In this implementation, each engine is connected to a dedicated DMA engine, with all DMA engines running asynchronously. In the DSE of \mbox{f-CN$\text{N}^{\text{x}}$}, the \textbf{FPSobj} objective function \mbox{(Eq. (\ref{eq:fpsobj}))} is employed, with a target frame rate of 25 fps for ZFNet, PilotNet and SceneLabelCNN, and \mbox{4 fps} for VGG16\footnote{By using $fps^{\text{target}}_i$$=$$ \min(fps^{\text{user}}_i,fps^{\text{max}}_i)$ as per Eq. (\ref{eq:fpsobj}), VGG16 achieves $fps^{\text{max}}_i$ of around 4 fps with the proposed architecture on ZC706.}.
	
	\vspace{-0.2cm}
	Table \ref{table:comp_base} shows the actual performance for each design as measured on the ZC706 board under varying bandwidth budget. 
	In bandwidth restricted cases (rows 1-3,5-7), f-CN$\text{N}^{\text{x}}$ outperforms the baseline by up to 77\% and 91\% in average throughput across the CNNs of each benchmark and with over 50\% improvement on the achieved \textbf{FPSobj} values. As more bandwidth becomes available \mbox{(rows 4, 8)}, the two accelerators become more compute-bound and the difference in performance tends to decrease. Due to the asynchronous operation of the contention-unaware accelerator's DMA engines, 
	different memory transactions affect each other and degrade the overall bandwidth utilisation efficiency, which in turn causes the CNN engines to remain underutilised. On the other hand, the \mbox{f-CN$\text{N}^{\text{x}}$} 
	alleviates the effect of randomised bandwidth contention between CNN engines and 
	sustains a high utilisation of the hardware, by using its memory-aware scheme 
	{\color{black}that couples the optimisation of the compute resources and the external memory bandwidth, and outperforms the contention-unaware in cases where bandwidth is the critical factor.}

	\begin{table*}[t]
		\vspace{-0.2cm}
		\centering
		\caption{Comparison of f-CN$\text{N}^{\text{x}}$ (ZC706) and GPU (NVIDIA TX1) on multi-CNN benchmarks (batch size = 1)}
		\vspace{-0.15cm}
		\label{gpu_comp_table}
		\footnotesize
		\resizebox{0.775\linewidth}{!}{
			\setlength\tabcolsep{2pt} 
			\begin{tabular}{c l l l | l l | l l | l l | l l}
				\toprule
				Benchmark & \multicolumn{1}{l}{Model Set} & \begin{tabular}[c]{@{}l@{}} f-CN$\text{N}^{\text{x}}$ \\ (GOp/s) \end{tabular} & \begin{tabular}[c]{@{}l@{}} f-CN$\text{N}^{\text{x}}$ \\ (GOp/s/W) \end{tabular} & \begin{tabular}[c]{@{}l@{}} TX1 \\ (GOp/s) \end{tabular} & \begin{tabular}[c]{@{}l@{}} TX1 \\ (GOp/s/W) \end{tabular} & \begin{tabular}[c]{@{}l@{}} TX1 (5W) \\ (GOp/s)  \end{tabular} & \begin{tabular}[c]{@{}l@{}} TX1 (5W) \\ (GOp/s/W)  \end{tabular} & \begin{tabular}[c]{@{}l@{}} Gain \\ (GOp/s) \end{tabular} & \begin{tabular}[c]{@{}l@{}} Gain \\ (GOp/s/W) \end{tabular} & \begin{tabular}[c]{@{}l@{}} Gain (5W) \\ (GOp/s) \end{tabular} & \begin{tabular}[c]{@{}l@{}} Gain (5W) \\ (GOp/s/W) \end{tabular} \\ 
				\midrule
				& ZFNet & 23.05 (10.37 fps) & \phantom{1}5.76 (2.59 fps/W) & \phantom{1}29.51 & \phantom{1}1.84 & \phantom{2}3.72 & \phantom{2}0.74 & 0.78$\times$ & 3.13$\times$ & \phantom{2}6.19$\times$ & 7.74$\times$ \\
				3 CNNs & SceneLabelCNN & 99.21 (12.96 fps) & 24.80 (3.24 fps/W) & 101.64 & \phantom{1}6.35 & 12.81 & \phantom{2}2.56 & 0.97$\times$ & 3.90$\times$ & \phantom{2}7.74$\times$ & 9.68$\times$ \\
				& VGG16 & 79.63 (\phantom{1}2.59 fps) & 19.90 (0.65 fps/W) & 408.03 & 25.50 & 51.43 & 10.28 & 0.19$\times$ & 0.78$\times$ & \phantom{2}1.55$\times$ & 1.93$\times$ \\
				\midrule
				& Average (geo. mean) & - & - & - & - & - & - & 0.53$\times$ & 2.12$\times$ & \phantom{2}4.20$\times$ & 5.25$\times$ \\
				\midrule
				\midrule
				& ZFNet & 16.28 (\phantom{2}7.32 fps) & \phantom{1}4.07 (1.83 fps/W) & \phantom{1}29.37 & \phantom{1}1.83 & \phantom{2}3.70 & \phantom{2}0.74 & 0.55$\times$ & 2.21$\times$ & \phantom{2}4.40$\times$ & \phantom{2}5.50$\times$ \\
				& PilotNet & \phantom{1}1.97 (31.77 fps) & \phantom{1}0.49 (7.94 fps/W) & \phantom{10}0.82 & \phantom{1}0.05 & \phantom{2}0.10 & \phantom{2}0.02 & 2.40$\times$ & 9.61$\times$ & 19.09$\times$ & 23.86$\times$ \\
				4 CNNs & SceneLabelCNN & 93.43 (12.21 fps) & 23.36 (3.05 fps/W) & 101.17 & \phantom{1}6.32 & 12.73 & \phantom{2}2.54 & 0.92$\times$ & 3.69$\times$ & \phantom{0}7.33$\times$ & \phantom{2}9.17$\times$ \\
				& VGG16 & 75.00 (\phantom{1}2.44 fps) & 18.75 (0.61 fps/W) & 406.15 & 25.38 & 51.12 & 10.22 & 0.18$\times$ & 0.74$\times$ & \phantom{0}1.46$\times$ & \phantom{2}1.83$\times$ \\
				
				\midrule
				& Average (geo. mean) & - & - & - & - & - & & 0.69$\times$ & 2.76$\times$ & \phantom{2}5.48$\times$ & \phantom{2}6.85$\times$ \\

				%
				\bottomrule
			\end{tabular}
		}
		\vspace{-0.55cm}
	\end{table*}
	
	\vspace{-0.425cm}
	\subsection{Comparison with Embedded GPU}
	\vspace{-0.15cm}
	With a large number of CNNs being deployed for inference in multi-tasking embedded systems, 
	our evaluation focuses on the embedded space. In power-constrained applications, the primary metrics of interest comprise 1) the absolute power consumption and 2) the performance efficiency in terms of performance-per-Watt. In this respect, we investigate the performance efficiency of f-CN$\text{N}^{\text{x}}$ on Zynq XC7Z045 in relation to the widely used NVIDIA Tegra X1 (TX1). To comply with the stringent latency needs of modern systems, both the FPGA and GPU designs use a batch size of 1. 
	
	\vspace{-0.2cm}
	For the performance evaluation on TX1, we use NVIDIA TensorRT as supplied by the JetPack 3.1 package. TensorRT is run with the cuDNN library and 16-bit half-precision floating-point arithmetic (FP16) which enables the highly optimised execution of layers. In each benchmark, the TensorRT implementations of the target CNNs are scheduled over the GPU in a rotational and periodic manner. Across all the platforms, each multi-CNN benchmark is run 100 times to obtain the average performance. Furthermore, power measurements for the GPU and the FPGA are obtained via a power monitor on the corresponding board. In all cases, we subtract the average idle power\footnote{Idle Power: Jetson  TX1 (5W), ZC706 (7W).} from the measurement to obtain the power due to benchmark execution which includes the off-chip memory. {\color{black}The idle power of the ZC706 platform is measured at the board level with no design programmed in the FPGA fabric, so that the clock tree power and the power leakage of the chip are also included in the run-time power due to benchmark execution.} 
	
\begin{figure}
		\vspace{-0.3cm}
		\centering
		
		\begin{subfigure}{.48\linewidth}
			\centering
			{\includegraphics[trim={1.9cm 8cm 2cm 8cm},clip,width=1\linewidth]{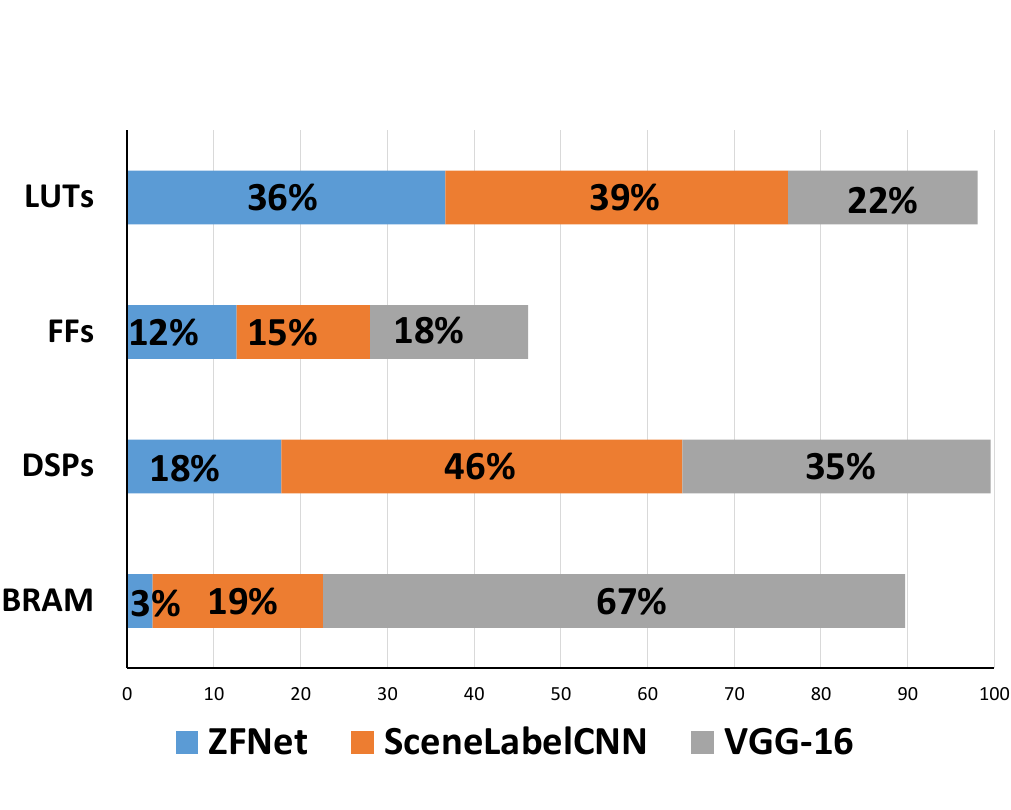}}
			\vspace{-0.6cm}
			\caption{ }
			\label{fig:rsc_util_3cnns}
		\end{subfigure}
		\begin{subfigure}{.48\linewidth}
			\centering
			{\includegraphics[trim={1.9cm 8cm 2cm 8cm},clip,width=1\linewidth]{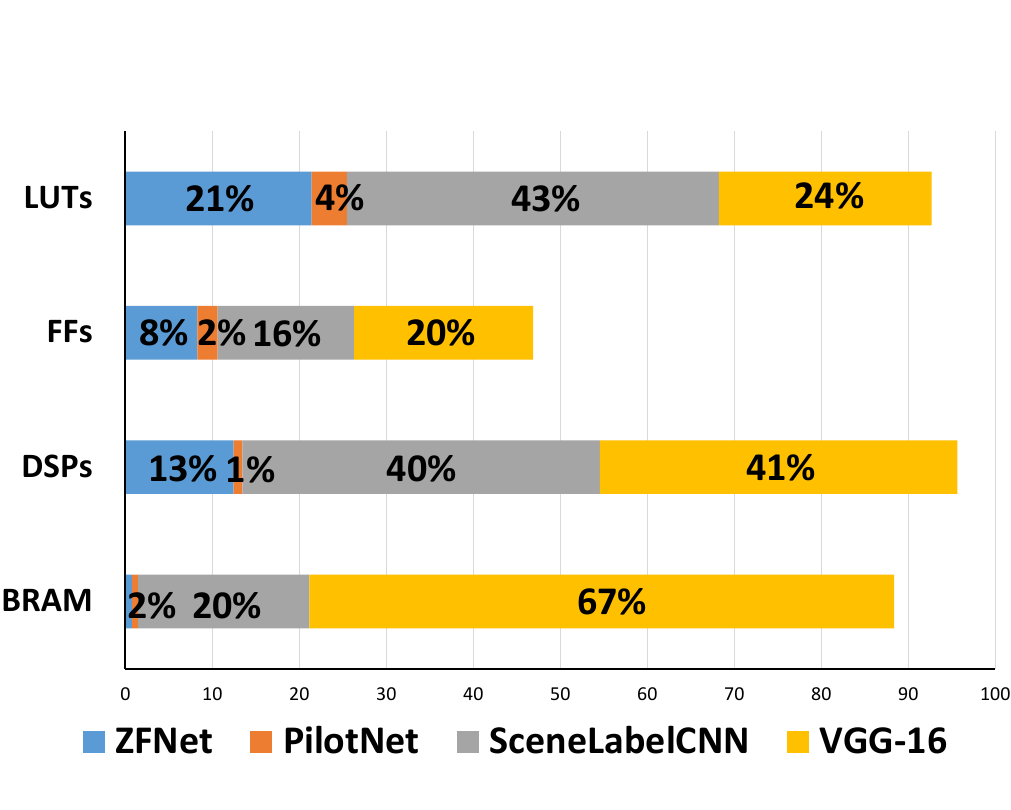}}
			\vspace{-0.6cm}
			\caption{ }
			\label{fig:rsc_util_4cnns}
		\end{subfigure}
		\vspace{-0.5cm}
		\caption{Resource utilisation breakdown of the f-CN$\text{N}^{\text{x}}$ designs for (\ref{fig:rsc_util_3cnns}) Table \ref{gpu_comp_table}:3 CNNs and (\ref{fig:rsc_util_4cnns}) Table \ref{gpu_comp_table}:4 CNNs.}
		\label{fig:rsc_util}
		\vspace{-0.3cm}
	\end{figure}

	\vspace{-0.2cm}
	\mbox{Table \ref{gpu_comp_table}} shows the measured performance efficiency of TX1 and ZC706. For all benchmarks, the target objective function was \textbf{FPSobj} (Eq. (\ref{eq:fpsobj})) with $fps^{\text{target}}_i$ set to 25 fps for ZFNet, PilotNet and SceneLabelCNN and 4 fps for VGG16. 
%
	TX1 mounts a 256-core GPU with hardware support for FP16 arithmetic and with a configurable range of frequencies up to 998 MHz at a peak power consumption of around 15W. 
	To investigate the performance of each platform under the same power constraints, we set a power budget of 5W {\color{black}as is commonly present in autonomous vehicles}, and configure the GPU with a 76.8 MHz clock rate in order not to exceed the 5W dynamic power budget.
	In the case of three CNNs, f-CN$\text{N}^{\text{x}}$ achieves a throughput improvement of up to 7.74$\times$ with an average of 4.2$\times$ (geo. mean) across the three models. In the case of four CNNs, f-CN$\text{N}^{\text{x}}$ demonstrates a throughput gain of up to 19.09$\times$ with an average of 5.48$\times$ (geo. mean) across the four models. Fig. \ref{fig:rsc_util} shows the post place-and-route resource utilisation breakdown between the CNN engines. 
	\mbox{f-CNN$^\text{x}$} allocates effectively a higher amount of FPGA resources for the more computationally heavy SceneLabelCNN and VGG16 to balance the achieved fps-per-CNN as dictated by \textbf{FPSobj}. The \texttt{MCNN-HS} module adds a minimal resource overhead of less than 5\% in LUTs and FFs, with the BRAMs of the staging buffers included and equally spread over the CNNs in Fig. \ref{fig:rsc_util}.
	
	\vspace{-0.2cm}
	To evaluate performance efficiency, we configure the GPU with the peak frequency of 998 MHz. When running the three CNNs, f-CN$\text{N}^{\text{x}}$ overpasses the performance-per-Watt of TX1 by up to \mbox{3.9$\times$ with} an average of \mbox{2.12$\times$ (}geo. mean). In the four-CNN benchmark, f-CN$\text{N}^{\text{x}}$ yields up to \mbox{9.61$\times$ gain} over TX1 in performance efficiency with an average of 2.76$\times$ (geo. mean) across the four CNNs. Despite the fact that the GPU executes CNN layers very efficiently, existing highly optimised implementations are limited to a sequential scheduling of layers and CNNs. 
	In contrast, f-CN$\text{N}^{\text{x}}$ exploits both the pipelined parallelism between layers within a CNN engine and the parallel execution of CNNs across multiple engines, and generates designs tailored to the target application.
	
	\section{Related Work}
	\label{sec:related_work}
	Since the original publication of this work, there have been several proposed hardware designs that specifically target multi-tenant multi-CNN applications. Closer to f-CNN$^\text{x}$, Kwon \textit{et al.}~\cite{hda2021hpca} introduce heterogeneous dataflow accelerators (HDAs), which employ multiple sub-accelerators, each supporting a different dataflow. Based on their workload characteristics, each CNN and each CNN layer can be scheduled at run time to the highest performing sub-accelerator. Planaria~\cite{planaria2020micro} proposes the dynamic construction of compute engines using multiple distinct systolic arrays. As such, the compute engine of the accelerator can be configured at run time and each DNN layer can be scheduled to the most suitable constructed compute engine at any time instance. AI-MT~\cite{aimt2020isca} schedules together compute- and memory-bound layers of different CNNs in order to fully utilise both the computational resources and the available off-chip memory bandwidth. Towards a CNN-hardware co-design approach, Yang \textit{et al.}~\cite{multidnn_codesign2020dac} propose a methodology to co-design multiple CNNs targeting different tasks, together with the underlying multi-CNN accelerator. Finally, Ribes \textit{et al.}~\cite{multilstms2020fpt} propose a multi-LSTM accelerator.
	
	Different to these works, f-CNN$^\text{x}$ adopts a more customisation-driven strategy by instantiating multiple \textit{spatially distinct} compute engines, each one tailored at a fine granularity for the particular CNN. Furthermore, f-CNN$^\text{x}$ co-optimises the design of these compute engines with the scheduling of the off-chip memory transfers to achieve high resource utilisation at run time. Overall, this approach allows us to reach maximum performance and energy efficiency, at the cost of limited programmability upon deployment.
	
	\vspace{-0.3cm}
	\section{Conclusion}
	\vspace{-0.1cm}
	This paper presents f-CNN$^\text{x}$, a framework for mapping multiple CNNs on FPGAs. By introducing a highly-customisable multi-CNN architecture together with an external memory access policy, the proposed toolflow tailors the allocation of both compute resources and external memory bandwidth to the performance requirements of the target set of CNNs.  
	Evaluation shows that f-CNN$^\text{x}$ achieves performance gains of up to 50\% over mappings that allow memory contention and delivers up to 6.8$\times$ higher performance-per-Watt over highly optimised embedded GPU designs. 
	To the best of our knowledge, this work introduces for the first time in the literature the mapping of multiple CNNs. {\color{black}Future work will explore the mapping of multiple CNN workloads in cloud \mbox{environments}.}

	\vspace{-0.3cm}
	
	\bibliographystyle{IEEEtran}
	\bibliography{Bibliography}

\begin{thebibliography}{10}
\providecommand{\url}[1]{#1}
\csname url@samestyle\endcsname
\providecommand{\newblock}{\relax}
\providecommand{\bibinfo}[2]{#2}
\providecommand{\BIBentrySTDinterwordspacing}{\spaceskip=0pt\relax}
\providecommand{\BIBentryALTinterwordstretchfactor}{4}
\providecommand{\BIBentryALTinterwordspacing}{\spaceskip=\fontdimen2\font plus
\BIBentryALTinterwordstretchfactor\fontdimen3\font minus
  \fontdimen4\font\relax}
\providecommand{\BIBforeignlanguage}[2]{{%
\expandafter\ifx\csname l@#1\endcsname\relax
\typeout{** WARNING: IEEEtran.bst: No hyphenation pattern has been}%
\typeout{** loaded for the language `#1'. Using the pattern for}%
\typeout{** the default language instead.}%
\else
\language=\csname l@#1\endcsname
\fi
#2}}
\providecommand{\BIBdecl}{\relax}
\BIBdecl

\bibitem{Chen2015}
C.~Chen, A.~Seff, A.~Kornhauser, and J.~Xiao, ``Deep{D}riving: Learning
  affordance for direct perception in autonomous driving,'' in \emph{ICCV},
  2015.

\bibitem{nvidia_2016}
M.~Bojarski \emph{et~al.}, ``{End to End Learning for Self-Driving Cars},''
  \emph{CoRR}, 2016.

\bibitem{Ren_2017}
S.~Ren, K.~He, R.~Girshick, and J.~Sun, ``Faster {R-CNN}: Towards real-time
  object detection with region proposal networks,'' \emph{TPAMI}, 2017.

\bibitem{Badrinarayanan_2017}
V.~Badrinarayanan \emph{et~al.}, ``{SegNet: A Deep Convolutional
  Encoder-Decoder Architecture for Scene Segmentation},'' \emph{TPAMI}, 2017.

\bibitem{Caulfield2016}
A.~M. Caulfield \emph{et~al.}, ``{A Cloud-Scale Acceleration Architecture},''
  in \emph{MICRO}, 2016.

\bibitem{Venieris_2017b}
S.~I. Venieris and C.-S. Bouganis, ``{Latency-Driven Design for FPGA-based
  Convolutional Neural Networks},'' in \emph{FPL}, 2017.

\bibitem{Yufei_Ma_2017}
Y.~Ma \emph{et~al.}, ``{An Automatic {RTL} Compiler for High-Throughput {FPGA}
  Implementation of Diverse Convolutional Neural Networks},'' in \emph{{FPL}},
  2017.

\bibitem{sv2018csur}
S.~I. Venieris, A.~Kouris, and C.-S. Bouganis, ``{Toolflows for Mapping
  Convolutional Neural Networks on FPGAs: {A} Survey and Future Directions},''
  \emph{ACM Computing Surveys}, 2018.

\bibitem{Smolyanskiy_2017}
N.~Smolyanskiy \emph{et~al.}, ``{Toward Low-Flying Autonomous MAV Trail
  Navigation using Deep Neural Networks for Environmental Awareness},'' in
  \emph{IROS}, 2017.

\bibitem{He_2016}
K.~He, X.~Zhang, S.~Ren, and J.~Sun, ``{Deep Residual Learning for Image
  Recognition},'' in \emph{{CVPR}}, 2016.

\bibitem{Howard2017mobilenet}
A.~G. Howard \emph{et~al.}, ``{MobileNets: Efficient convolutional neural
  networks for mobile vision applications},'' \emph{CoRR}, 2017.

\bibitem{Zoph2018cvpr}
B.~Zoph, V.~Vasudevan, J.~Shlens, and Q.~V. Le, ``{Learning Transferable
  Architectures for Scalable Image Recognition},'' in \emph{CVPR}, 2018.

\bibitem{Lee_1987}
E.~A. Lee \emph{et~al.}, ``{Synchronous Data Flow},'' \emph{Proc. of IEEE},
  1987.

\bibitem{Venieris_2016}
S.~I. Venieris and C.-S. Bouganis, ``{fpgaConvNet: A Framework for Mapping
  Convolutional Neural Networks on FPGAs},'' in \emph{{FCCM}}, 2016.

\bibitem{Draper1999}
D.~L. Draper, A.~K. Jonsson, D.~P. Clements, and D.~E. Joslin, ``{Cyclic
  Scheduling},'' in \emph{IJCAI}, 1999, pp. 1016--1021.

\bibitem{LEVNER2010}
E.~Levner, V.~Kats, D.~A.~L. de~Pablo, and T.~Cheng, ``{Complexity of Cyclic
  Scheduling Problems: A State-of-the-art Survey},'' \emph{CAIE}, 2010.

\bibitem{Micheli1994}
G.~D. Micheli, \emph{{Synthesis and Optimization of Digital Circuits}},
  1st~ed.\hskip 1em plus 0.5em minus 0.4em\relax McGraw-Hill Higher Education,
  1994.

\bibitem{Audet_2002}
C.~Audet and J.~J.~E.~Dennis, ``{Analysis of Generalized Pattern Searches},''
  \emph{SIAM Journal on Optimization}, 2002.

\bibitem{Lecun1998}
Y.~LeCun, L.~Bottou, Y.~Bengio, and P.~Haffner, ``{Gradient-Based Learning
  Applied to Document Recognition},'' in \emph{Proc. of IEEE}, 1998.

\bibitem{Krizhevsky09learningmultiple}
A.~Krizhevsky, ``{Learning Multiple Layers of Features from Tiny Images},''
  University of Toronto, Tech. Rep., 2009.

\bibitem{Zeiler_2014}
M.~D. Zeiler and R.~Fergus, ``{Visualizing and Understanding Convolutional
  Networks},'' in \emph{{ECCV}}, 2014.

\bibitem{Cavigelli2015}
L.~Cavigelli, M.~Magno, and L.~Benini, ``{Accelerating real-time embedded scene
  labeling with convolutional networks},'' in \emph{DAC}, 2015.

\bibitem{Simonyan14c}
K.~Simonyan and A.~Zisserman, ``{Very Deep Convolutional Networks for
  Large-Scale Image Recognition},'' \emph{ICLR}, 2015.

\bibitem{hda2021hpca}
H.~Kwon, L.~Lai, M.~Pellauer, T.~Krishna, Y.-H. Chen, and V.~Chandra,
  ``{Heterogeneous Dataflow Accelerators for Multi-DNN Workloads},'' in
  \emph{HPCA}, 2021.

\bibitem{planaria2020micro}
S.~Ghodrati, B.~H. Ahn, J.~Kyung~Kim, S.~Kinzer, B.~R. Yatham, N.~Alla,
  H.~Sharma, M.~Alian, E.~Ebrahimi, N.~S. Kim, C.~Young, and H.~Esmaeilzadeh,
  ``{Planaria: Dynamic Architecture Fission for Spatial Multi-Tenant
  Acceleration of Deep Neural Networks},'' in \emph{MICRO}, 2020.

\bibitem{aimt2020isca}
E.~Baek, D.~Kwon, and J.~Kim, ``{A Multi-Neural Network Acceleration
  Architecture},'' in \emph{ISCA}, 2020.

\bibitem{multidnn_codesign2020dac}
L.~Yang, Z.~Yan, M.~Li, H.~Kwon, W.~Jiang, L.~Lai, Y.~Shi, T.~Krishna, and
  V.~Chandra, ``{Co-Exploration of Neural Architectures and Heterogeneous ASIC
  Accelerator Designs Targeting Multiple Tasks},'' in \emph{DAC}, 2020.

\bibitem{multilstms2020fpt}
S.~Ribes, P.~Trancoso, I.~Sourdis, and C.-S. Bouganis, ``{Mapping Multiple LSTM
  models on FPGAs},'' in \emph{ICFPT}, 2020.

\end{thebibliography}
	
\end{document}